# iMedBot: A Web-based Intelligent Agent for Healthcare Related Prediction and Deep Learning

Chuhan Xu, Xia Jiang

**Abstract:**

**Background:** Breast cancer is a multifactorial disease, genetic and environmental factors will affect its incidence probability. Breast cancer metastasis is one of the main cause of breast cancer related deaths reported by the American Cancer Society (ACS). **Method: the** iMedBot is a web application that we developed using the python Flask web framework and deployed on Amazon Web Services. It contains a frontend and a backend. The backend is supported by a python program we developed using the python Keras and scikit-learn packages, which can be used to learn deep feedforward neural network (DFNN) models. **Result:** the iMedBot can provide two main services: 1. it can predict 5-, 10-, or 15-year breast cancer metastasis based on a set of clinical information provided by a user. The prediction is done by using a set of DFNN models that were pretrained, and 2. It can train DFNN models for a user using user-provided dataset. The model trained will be evaluated using AUC and both the AUC value and the AUC ROC curve will be provided. **Conclusion:** The iMedBot web application provides a user-friendly interface for user-agent interaction in conducting personalized prediction and model training. It is an initial attempt to convert results of deep learning research into an online tool that may stir further research interests in this direction.

**Keywords:** Deep learning, Breast Cancer, Web application, Model training.

## 1. Introduction

This paper focuses on an introduction of the iMedBot – A web-based Intelligent Agent, which was initially developed as an user friendly and interactive online agent for predicting n-year breast cancer metastasis. The use of the current version of iMedBot is limited to the research community for the purpose of boosting the deployment and dissemination of research results concerning deep learning, machine learning, and other methods of artificial intelligence (AI), further stirring research interests in AI in medicine, and setting an example of a web-based intelligent agent that can assist medical activities such as prognosis and decision support. As shown in Figure 1, the iMedBot is a full-stack web application that consists of both a front end a back end. The back end contains the model training service module, the model prediction service model, the serialized model object, and the KerasClassifier module. The later two S3 isare the supporting hardware components developed by using the resources provided by the Amazon web services (AWS). The current core services provided by the back end include 1) Model training service; and 2) personalized prediction of 5-, 10-, and 15-year breast cancer metastasis. The front end of the iMedBot consists of the client, the domain name server (DNS), and a load Balancer. The route53, ELB, EC2later two are also developed usiAWS services resources to support the main functions of the front end. The client module provides the agent-user interface, which supports the agent-user interaction by a set of "conversation" windows. The main window hosts the sequence of dialogues once the "conversation" between the again and user begins.

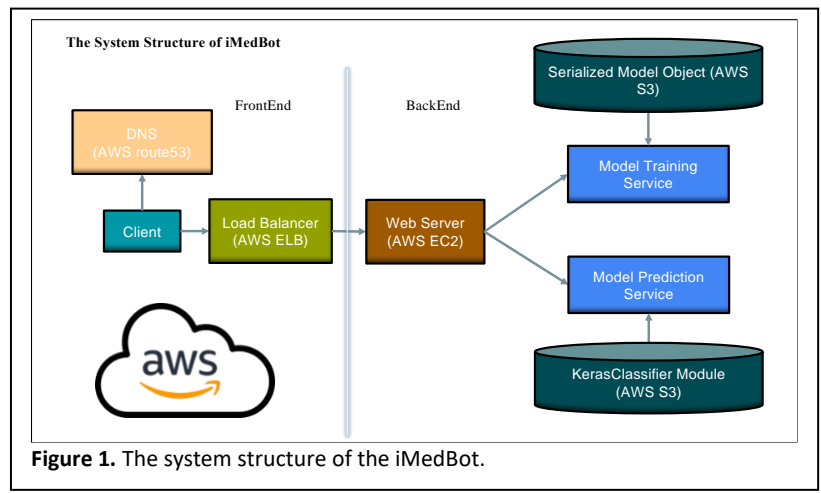

**Figure 1.** The system structure of the iMedBot.

## 2. The Backend Services

The model training service is provided by our python deep feedforward neural network (DFNN) programs [1], [2], which were developed by using the Keras python package [3]. We included this service in the iMedBot because we assume there are scenarios in which the users prefer training prediction models using their own data. When a model training service call is initialized in the front end and passed to the back end, the dataset provided by the user will be split in stratified manner: 80% of the data will be used to train models following the 5-fold cross validation strategies and 20% of the data will be used as the validation dataset. The validation AUC is one of the output of the model training service, which will be displayed by the [4].which will be generated by testing the best output model of a grid search using the validation dataset. Grid search is a systematic way of conduct hyperparameter tuning to identify the best performing models in deep learning [3]. Our DFNN models have 13 hyperparameters. We call one particular value assignment of the set of hyperparameters of our DFNN models a hyperparameter setting. So each DFNN model has a unique hyperparameter setting. Before a grid search, each of the hyperparameters is given a range of values the hyperparameter can take. Grid search will train all DFNN models corresponding to all possible hyperparameter settings determined by the preselected ranges of values for the hyperparameters. The results returned by the model training service include the validation AUC, the ROC curve plot [4], and the best prediction model found by grid search.

The prediction service is provided by a set of deep feedforward neural network (DFNN) models that were pretrained with our DFNN programs [3] by Grid search strategy.[3] Grid search will train all DFNN models corresponding to all possible hyperparameter settings determined by the preselected ranges of values for the hyperparameters. The same procedures used in model training service were used for pretraining these models. The difference between these two kinds of services is that the model training service is meant to train models using user-provided datasets, while the pretrained DFNN models were trained using the LSM 5-, 10-, 15-year datasets that are publicly available [3], [5]. The pretrained DFNN models are the best performing models selected from a large set of models trained via grid search. A detailed description of the DFNN model training, grid search, and model evaluation is provided in [3], [6]. The pretrained DFNN models contain a set of predictors [3], which are the clinical features defined by the datasets from which these models were trained; These models can be used to conduct personalized prediction once the patient-specific values of the predictors are received.

In addition to the two core services, the back end contains other python tools that we developed for tasks such as processing input data, analyzing results, and evaluating the prediction performance of models. These tools provide some of the assistant services such as generating the ROC curves based on the 5-fold cross validation to compare the prediction performance of different models [3]. The iMedBot is currently hosted at the AWS (Amazon Web Services), and therefore there are other supporting system components in the backend, which are provided by AWS S3.

## 3. The Front End

The front end of the iMedBot has a main user friendly agent-user interaction window, which hosts the sequence of dialogues once the "conversion" between the again and user begins. The possible dialogues are designed based on the current two core back end services that the iMedBot can provide: the prediction and model training. For example, to provide the personalized prediction service, the iMedBot will initialize a sequence of dialogues to go through with the user the set of the predictors of the back end DFNN models one after another. For each of the predictors, the iMedBot will provide to the user a list of all possible values of the predictor, from which ana user can select his/her patient specific value with an action as simple as clicking a button. Once the iMedBot receives the user input/responses, it will communicate with the proper back end components, pass over the user input collected through the "conversation" to the best model object that we deployed in our S3 bucket in the backend, and then receive and display the results from the back end. Figure 2, 3 show two examples of these agent-user interaction dialogues. The first example shows the dialogue with which the iMedBot prompts a user to enter (select) the specific value of the user's patient for a predictor called DCIS_ level(type of ductal carcinoma in situ), when the prediction service is called by the user. Towards the end of the prediction service call, the iMedBot will ask the users to do a simple survey about their experience with the prediction service call.

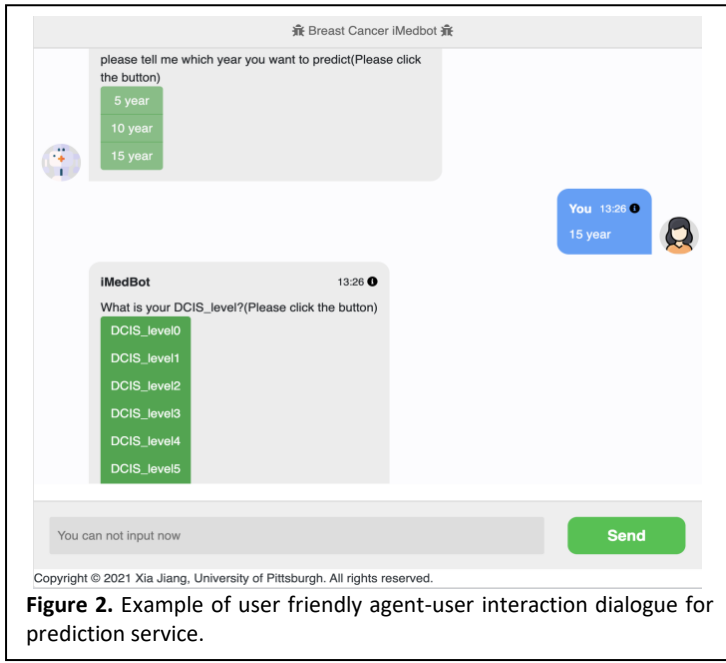

**Figure 2.** Example of user friendly agent-user interaction dialogue for prediction service.

To provide the model training service, the iMedBot will initialize the dialogues for user dataset upload, conduct error checking of the dataset, and allow the user to review the dataset once it is uploaded. The iMedBot also prompt a user to enter their preferred hyperparameter settings that can be tuned via grid search to improve the prediction performance of the output models. The next step is to set the hyperparameters for the deep learning model including learning rate, epochs, batchsize and so on, we also provide default settings. Once users finish uploading dataset and hyperparameter setting assignment, the backend model training service will be called. The output of the service call, currently including the validation AUC, a ROC curve plot, and the prediction model will be returned to the frontend and displayed in a dialogue. Then, the current model training service call ends, and the iMedBot will prompt the user for more actions that the user can take2. The second example in Figure 3 shows one of the output display dialog towards the end of a model training service call, in which the iMedBot presents to the user the validation AUC and ROC curve for the model trained using the user provided data, when the model training service is called by the user. The usage case flow charts for both the prediction and model training service calls in the front end are shown in Figure 4 below.

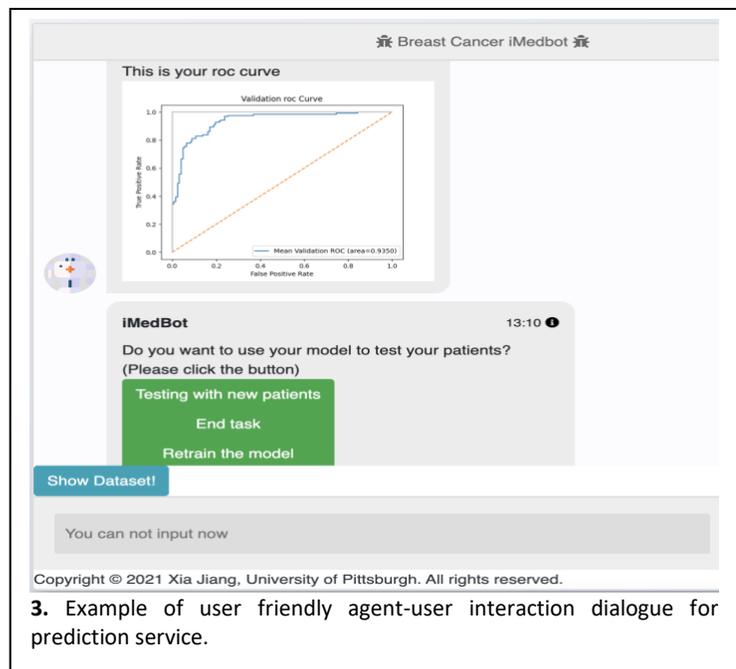

**3.** Example of user friendly agent-user interaction dialogue for prediction service.

4. **The main software development and storage tools/services**

Flask is a lightweight customizable framework written in Python [7]. The reason we chose Flask as our main framework is because we also used the Keras python package [8] to develop our DFNN programs, and the consistency of the language can help to reduce errors and make development and deployment more compatible. We mainly used HTML, CSS, JavaScript for front-end development and python for the backend. For the model prediction part, we applied model serialization technology to save our model to h5 format [9]. Github [10] provides free storage and version control services of our software. Our local development work is pushed to github [10] on daily basis, and any update of the work will be deployed automatically via a pipeline to the AWS platform provided by the paid Elastic Beanstalk Service [11]. The pipeline is also provided by a paid AWS service called [12]. As previously described our FDNN programs and other service tools are mainly developed in python using exiting python packages such as the Keras python package provided in scikit-learn [13] and the matplotlib package [14] for generating figures.

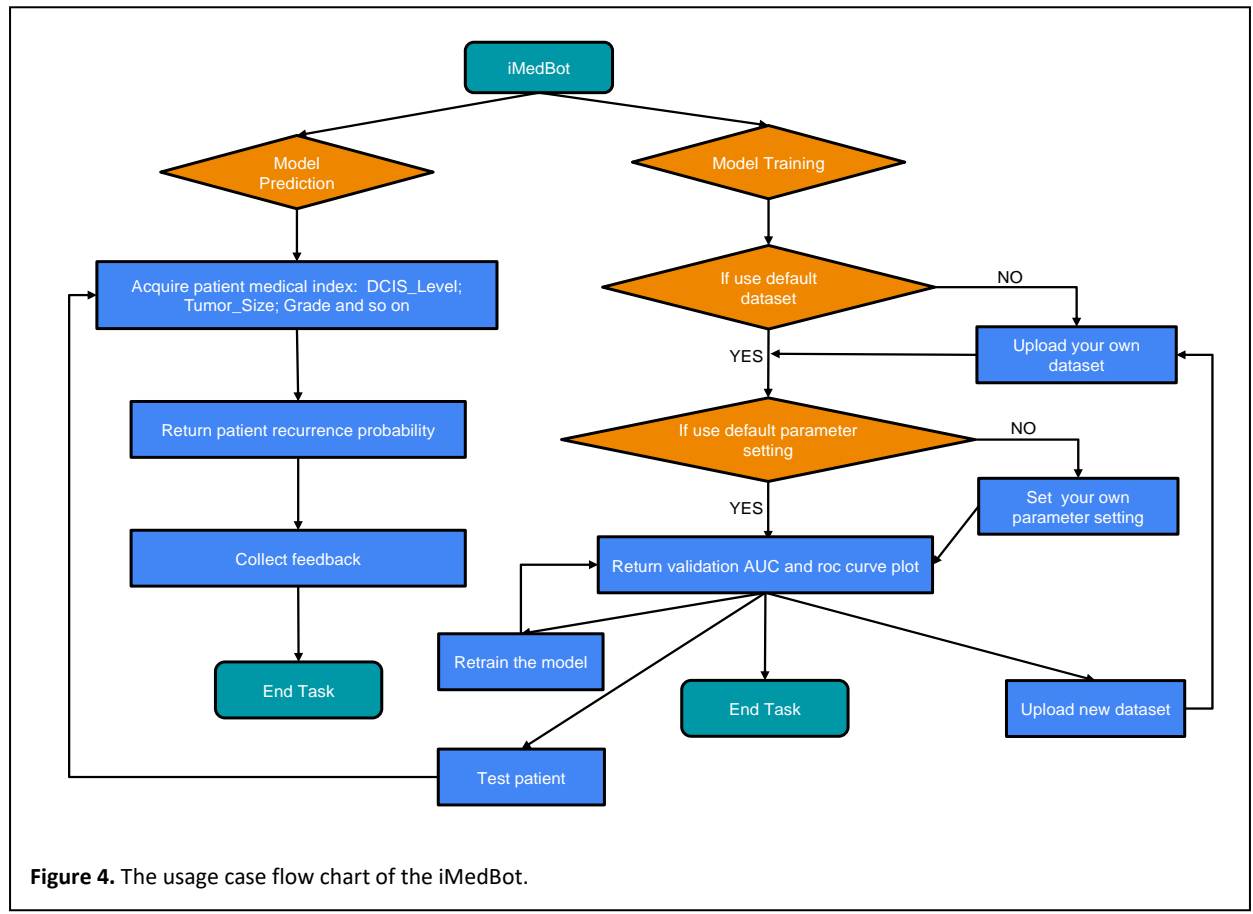

**Figure 4.** The usage case flow chart of the iMedBot.

## 5. The main supporting hardware components

Figure 1 also illustrates the main supporting hardware components of the iMedBot, provided by AWS. The four main components include DNS, Load Balancer, Web Server, and model objects. AWS Route53 is a scalable and highly available Domain Name System service (DNS) which provides simple and short URL to help our clients easily get access to the iMedBot web application [15]. AWS Load Balancer is used to automatically distributes clients' incoming traffic across multiple targets to decrease the risk of break down when a lot of users access the iMedBot on the same time. The important component is AWS EC2 where we put our source code in, EC2 is the AWS computing service, which offers computing capacity on demand [16]. Because the iMedBot needs to support the deep learning model training service which requires the EC2 instance has the powerful ability of CPU and memory, we changed our EC2 instance to medium type. The trained model will be saved into AWS S3 bucket, for the model prediction service, the model will be the best model based on breast cancer LSM dataset trained by ourselves. For the model training service, the model trained by users will be saveds as serialized model object, both of these two kinds of model will be stored in AWS S3 bucket [17].

## 6. Limitations

The reason we used Keras module is that all the data in the LSM (LSDS for Metastasis) dataset are category data [17]. According to the grid search experiment results, the current validation AUC of the best model is 84.3% according to our grid search experiment results, but we are continue doing more experiments now to further improve the model accuracy in the future. We have strict limitation of the dataset size which requires that the user can't uploaded dataset whose size is more than 500kb, because we need to make sure our current AWS EC2 instance can support the training computing work, and the dataset must be category dataset, maybe we can deploy more training methods in the backend in addition to only supporting one KerasClassifier module.

## 7. Conclusion

So far, the iMedBot application can be accessed by using URL *http://iMedBot.odpac.net/*. The iMedBot web application provides a user-friendly interface for user-agent interaction in conducting personalized prediction and model training. It is an initial attempt to convert results of deep learning research into an online tool that may stir further research interests in this direction. We plan on further expanding the backend of iMedBot to include other services such as risk factor learning (both single and interactive), causing learning, and clinical decision support.

**Authors' Contribution:** All authors contributed to the preparation and revision of the manuscript.

**Funding:** Research reported in this paper was supported by the U.S. Department of Defense through the Breast Cancer Research Program under Award No. W81XWH1910495 (to XJ). Other than supplying funds, the funding agencies played no role in the research.

**Ethics approval and consent to participate:** Not applicable.

**Conflicts of Interest:** The authors declare no conflict of interest.